\pdfoutput=1

\documentclass[11pt]{article}

\usepackage[]{acl}

\usepackage{times}
\usepackage{latexsym}

\usepackage[T1]{fontenc}

\usepackage[utf8]{inputenc}

\usepackage{microtype}

\usepackage{amsmath}
\usepackage{amssymb}
\usepackage{color}
\usepackage{graphicx}
\usepackage{hyperref}
\usepackage{booktabs}
\usepackage[T1]{fontenc}
\usepackage{fancyvrb}
\usepackage{multirow}
\usepackage{array}
\usepackage{arydshln}
\usepackage{xcolor}
\usepackage{colortbl}
\usepackage{ulem}
\usepackage{pifont}
\usepackage{stfloats}

\definecolor{right}{RGB}{0,128,96}
\definecolor{wrong}{RGB}{192,0,32}

\DeclareMathOperator*{\argmax}{arg\,max}

\newcolumntype{u}{>{\columncolor[RGB]{255, 248, 248}}c}
\newcolumntype{v}{>{\columncolor[RGB]{248, 248, 255}}c}
\newcolumntype{w}{>{\columncolor[RGB]{240, 240, 255}}c}
\newcolumntype{x}{>{\columncolor[RGB]{255, 255, 240}}c}
\newcolumntype{y}{>{\columncolor[RGB]{248, 255, 248}}c}
\newcolumntype{z}{>{\columncolor[RGB]{240, 255, 240}}c}

\usepackage{todonotes}
\newcounter{alisa}
\title{Generated Knowledge Prompting for Commonsense Reasoning}

\newcommand{\aspace}{\hspace{1em}}
\newcommand{\uw}{$^{\heartsuit}$}
\newcommand{\aitwo}{$^{\spadesuit}$}
\author{
    Jiacheng Liu\uw \aspace
    Alisa Liu\uw \aspace
    Ximing Lu\uw \aitwo \aspace
    Sean Welleck\uw \aitwo \aspace \\
\textbf{
    Peter West\uw \aitwo \aspace
    Ronan Le Bras\aitwo \aspace
    Yejin Choi\uw \aitwo \aspace
    Hannaneh Hajishirzi\uw \aitwo \aspace} \\
\uw Paul G. Allen School of Computer Science \& Engineering, University of Washington \\
\aitwo Allen Institute for Artificial Intelligence \\
\texttt{liujc@cs.washington.edu}
}

\begin{document}

\maketitle

\begin{abstract}
It remains an open question whether incorporating external knowledge benefits commonsense reasoning while maintaining the flexibility of pretrained sequence models.
To investigate this question, we develop generated knowledge prompting, which consists of generating knowledge from a language model, then providing the knowledge as additional input when answering a question.
Our method does not require task-specific supervision for knowledge integration, or access to a structured knowledge base, yet it improves performance of large-scale, state-of-the-art models
on four commonsense reasoning tasks, achieving state-of-the-art results on numerical commonsense (NumerSense), general commonsense (CommonsenseQA 2.0), and scientific commonsense (QASC) benchmarks.
Generated knowledge prompting 
highlights large-scale language models as flexible sources of external knowledge for improving commonsense reasoning.
Our code is available at \url{github.com/liujch1998/GKP}
\end{abstract}

\section{Introduction}
\label{sec:introduction}

\begin{figure}[t]
\centering
\includegraphics[width=\linewidth]{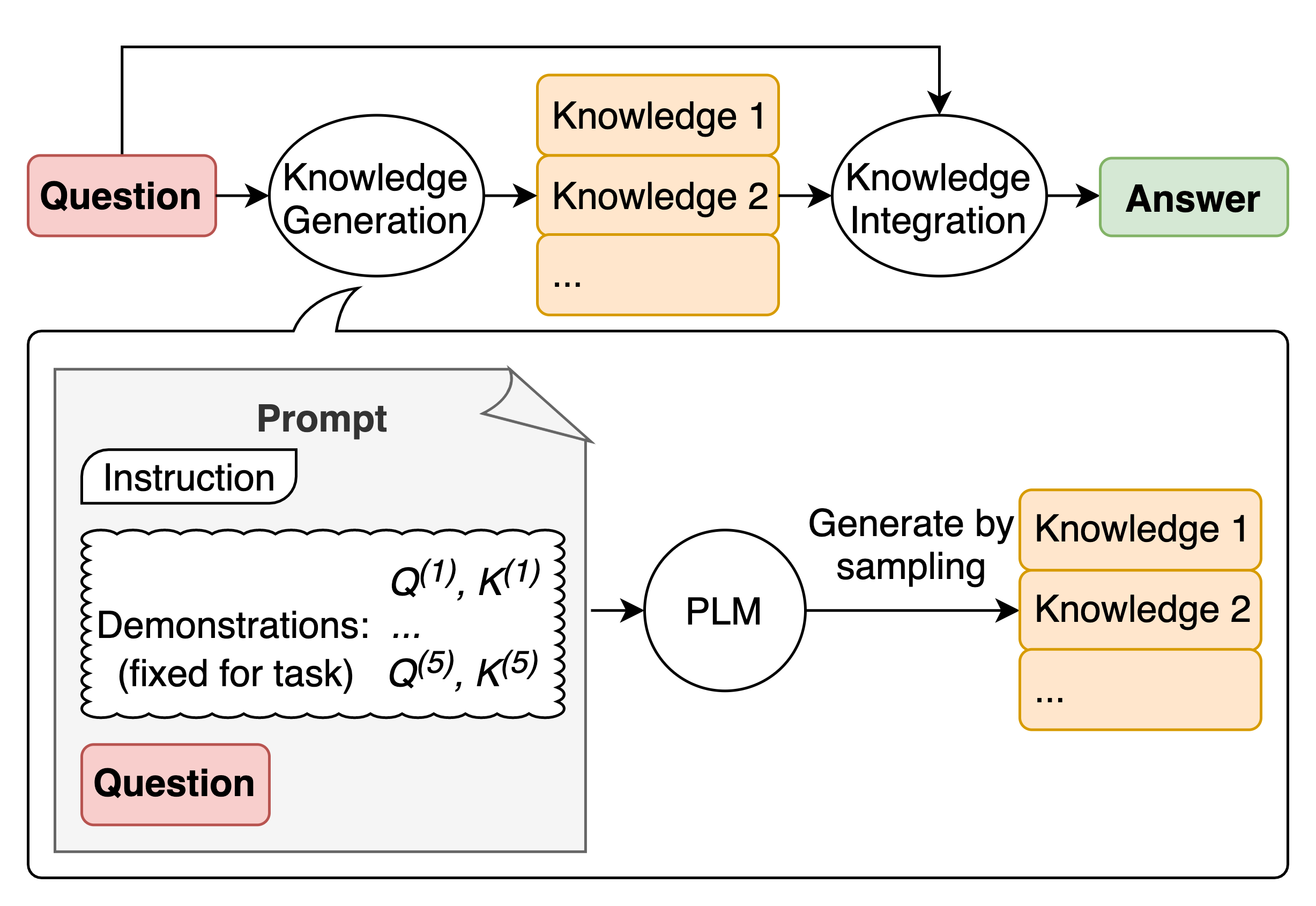}
\caption{
    Generated knowledge prompting involves (i) using few-shot demonstrations to generate question-related knowledge statements from a language model; 
    (ii) using a second language model to make predictions with each knowledge statement, then selecting the highest-confidence prediction.
}
\label{fig:diagram}
\end{figure}
\begin{table*}[t]
\footnotesize
\centering
\begin{tabular}{cp{9cm}cc}
\toprule
\textbf{Dataset} & \textbf{Question} / \textsl{Knowledge} & \textbf{Prediction} & \textbf{Score} \\
\midrule
\multirow{2}{*}{NumerSense} & \textbf{the word children means [M] or more kids.} & \textcolor{wrong}{one} & \textcolor{wrong}{0.37} | \textcolor{right}{0.35} \\
& \qquad \textsl{The word child means one kid.} & \textcolor{right}{two} & \textcolor{right}{0.91} \\
\midrule
\multirow{2}{*}{CSQA} & \textbf{She was always helping at the senior center, it brought her what?} & \textcolor{wrong}{feel better} & \textcolor{wrong}{0.97} | \textcolor{right}{0.02} \\
& \qquad \textsl{People who help others are usually happier.} & \textcolor{right}{happiness} & \textcolor{right}{0.98} \\
\midrule
\multirow{2}{*}{CSQA2} & \textbf{Part of golf is trying to get a higher point total than others.} & \textcolor{wrong}{yes} & \textcolor{wrong}{1.00} | \textcolor{right}{0.00} \\
& \qquad \textit{The player with the lowest score wins.} & \textcolor{right}{no} & \textcolor{right}{1.00} \\
\midrule
\multirow{2}{*}{QASC} & \textbf{Sponges eat primarily} & \textcolor{wrong}{cartilage} & \textcolor{wrong}{0.95} | \textcolor{right}{0.00} \\
& \qquad \textit{Sponges eat bacteria and other tiny organisms.} & \textcolor{right}{krill and plankton} & \textcolor{right}{0.99} \\
\bottomrule
\end{tabular}
\caption{
    Examples where prompting with generated knowledge rectifies model prediction.
    Each section shows \textcolor{right}{the correct answer in green}, \textcolor{wrong}{the incorrect answer in red}, and the prediction scores from the inference model that only sees the question (top) and the same model that sees the question prompted with the given knowledge (bottom).
}
\label{tab:examples}
\end{table*}

It remains an open research question whether external knowledge is needed for commonsense reasoning. On one hand, a substantial body of prior work has reported that integrating external knowledge can help improve task performance \cite[\textit{inter alia}]{mitra2019additional, bian2021benchmarking}, especially if the knowledge is high quality (e.g. hand-crafted by experts). On the other hand, recent leaderboards are often dominated by large-scale pretrained models that are fine-tuned on a target benchmark \cite{khashabi-etal-2020-unifiedqa, lourie2021unicorn}, suggesting that the benefits of external knowledge may wash away as the underlying models increase in size and are pretrained on ever larger amounts of raw text. 

Even if external knowledge is found to be effective on a particular task, \textit{flexibility} remains a fundamental hurdle to integrating external knowledge,
as many benchmarks currently lack appropriate knowledge bases with sufficient coverage. Furthermore, prior methods often require task-specific, custom supervision for knowledge integration \cite{mitra2019additional, chang-etal-2020-incorporating}, introducing a burden for rapidly adapting new pretrained models to a wide variety of tasks.

In this paper, we investigate whether external knowledge can be helpful for commonsense reasoning, even on top of the largest state-of-the-art pretrained models (e.g. T5-11b \cite{raffel2019exploring} and its variants), with a focus on four recent commonsense benchmarks.
To facilitate easier adaptation with any zero-shot or finetuned models, we propose an approach that does not require access to a structured knowledge base or joint finetuning for knowledge integration.

The key insight behind our method, Generated Knowledge Prompting (sketched in \autoref{fig:diagram}), is that we can generate useful knowledge from a language model, then provide the knowledge as an input prompt that is concatenated with a question.
%
To support a variety of settings without finetuning, the quality and flexibility of knowledge is crucial. We propose a simple, yet effective, method that elicits \textit{knowledge statements} (i.e. knowledge expressed as natural language statements) from generic language models in a few-shot setting. Compared to prior work that elicits knowledge via clarification questions \cite{shwartz-etal-2020-unsupervised} or contrastive explanations \cite{paranjape-etal-2021-prompting}, our approach can generate knowledge flexibly, beyond the scope of pre-defined templates (\autoref{tab:examples}). 


Experiments show that our method improves both zero-shot and finetuned models on numerical commonsense (NumerSense \cite{lin-etal-2020-birds}), general commonsense (CommonsenseQA \cite{talmor-etal-2019-commonsenseqa}, CommonsenseQA 2.0 \cite{talmor2021commonsenseqa}), and scientific commonsense (QASC \cite{khot2020qasc}) benchmarks, setting a new state-of-the-art on three of these datasets.
It outperforms the template-based knowledge generation method \textit{self-talk} \cite{shwartz-etal-2020-unsupervised}, while performing comparably to retrieval-based systems.

We find three factors contribute to the performance of generated knowledge prompting: (i) the \textit{quality} of knowledge, (ii) the \textit{quantity} of knowledge where the performance improves with more knowledge statements, and (iii) the strategy for integrating knowledge during inference. 
Our qualitative analysis suggests that the generated knowledge statements cover a variety of types, and can transform commonsense question answering to explicit reasoning procedures, e.g. deduction, that are supported by off-the-shelf and finetuned language models.

\section{Generated Knowledge Prompting}
\label{sec:method}

\begin{table*}[t]
\footnotesize
\centering
\resizebox{\textwidth}{!}{
\begin{tabular}{rp{7.5cm}p{8cm}}
\textbf{Task} & \textbf{NumerSense} & \textbf{QASC} \\
\midrule
\textbf{Prompt} & Generate some numerical facts about objects. Examples: & Generate some knowledge about the input. Examples: \\
\addlinespace[4pt]
& Input: \textbf{penguins have <mask> wings.} & Input: \textbf{What type of water formation is formed by clouds?} \\
& Knowledge: \textit{Birds have two wings. Penguin is a kind of bird.} & Knowledge: \textit{Clouds are made of water vapor.} \\
\addlinespace[4pt]
& ... & ... \\
\addlinespace[4pt]
& Input: \textbf{a typical human being has <mask> limbs.} & Input: \textbf{The process by which genes are passed is} \\
& Knowledge: \textit{Human has two arms and two legs.} & Knowledge: \textit{Genes are passed from parent to offspring.} \\
\addlinespace[4pt]
& Input: \textbf{\{question\}} & Input: \textbf{\{question\}} \\
& Knowledge: & Knowledge:
\end{tabular}
}
\caption{
    Prompts for knowledge generation for two of our tasks, NumerSense and QASC. 
    The prompt consists of an instruction, five demonstrations of question-knowledge pairs, and a new question placeholder.
    For full prompts on all the tasks we evaluate on, see Appendix \ref{sec:prompts}.
}
\label{tab:prompt}
\end{table*}

A multiple-choice commonsense reasoning task involves predicting an answer $a\in A_q$ given a question $q\in Q$, where the set of choices $A_q$ is finite and can vary by question, and both questions and answers are variable-length text sequences.
Our method answers commonsense questions in two steps.

The first step is \textit{knowledge generation}, where we use a language model $p_G(k|q)$ to generate knowledge statements conditioned on the question:
\begin{align*}
K_q &= \{ k_m : k_m \sim p_G(k|q), m = 1 \ldots M \},
\end{align*}
where each knowledge statement $k_m$ is a variable-length text sequence.
Intuitively, each statement contains information that is helpful for answering the question (e.g. \autoref{tab:examples}).

The second step is \textit{knowledge integration}, where generated knowledge is integrated into the decision process of a language model used for inference,
\begin{align*}
\hat{a} &= \argmax_{a \in A_q}{p_I(a|q,K_q)}.
\end{align*}
In contrast, the \textit{vanilla} setting of using the inference model without knowledge is represented by $\hat{a} = \argmax_{a \in A_q}{p_I(a|q)}$.


Next, we describe the knowledge generation and integration steps in detail.

\subsection{Knowledge Generation}
\label{sec:method_generation}

We generate question-related knowledge statements by prompting a language model.
The prompt consists of an instruction, a few demonstrations that are fixed for each task, and a new-question placeholder.
The demonstrations are human-written, and each consists of a question in the style of the task and a knowledge statement that is helpful for answering this question.
For a given task, we write five demonstrations using the format in \autoref{tab:prompt}.
%
%

We write questions (or select them from the training set, when available) that are representative of challenges posed by the task (e.g. numerical commonsense, scientific commonsense).
We pair each question with a knowledge statement that turns the commonsense problem posed by the question into an explicit reasoning procedure, without directly answering the question.
For example, the knowledge statement \textit{\uline{Birds have two wings. Penguin is a kind of bird.}} is helpful for the question \textbf{\uline{Penguins have <mask> wings}}, because it turns the problem into deductive reasoning.
Meanwhile, \textit{\uline{Penguins have two wings.}} would be a poor knowledge statement to demonstrate according to our guideline.

When generating knowledge for a new question $q$, we plug the question into the placeholder, and repeatedly sample generated continuations of this prompt to obtain a set of knowledge statements $K_q = \{ k_1, k_2, \hdots, k_M \}$.
For full prompts on all the tasks we evaluate on, see Appendix \ref{sec:prompts}. 

\subsection{Knowledge Integration via Prompting} 
\label{sec:method_inference}
In the knowledge integration step, we use a language model -- called the inference model -- to make predictions with each generated knowledge statement, then select the highest-confidence prediction.
Specifically, we use each knowledge statement to prompt the model, forming $M$ knowledge-augmented questions:
\begin{align*}
q_0 = q, q_1 = [k_1 || q], \hdots, q_M = [k_M || q],
\end{align*}
where $[\cdot || \cdot]$ denotes text concatenation.

We compute an aggregated score for each answer choice $a$ using the augmented question that best supports it under the inference model: 
\begin{align}
p_I(a|q,K_q) &\propto \max_{0 \le m \le M}{p_I(a|q_m)}. \label{eqn:max_ensembling}
\end{align}
Intuitively, this favors knowledge statements that strongly support one of the choices.

The predicted answer is then,
\begin{align*}
\hat{a} &= \argmax_{a \in A_q}{\max_{0 \le m \le M}{p_I(a|q_m)}},
\end{align*}
which is the choice that gets most support from one of the knowledge statements.
This prediction uses a single knowledge statement, which we refer to as the \textit{selected knowledge}: 
\begin{align*}
\hat{k} &= k_{\hat{m}} \text{ where } \hat{m} = \argmax_{0 \le m \le M}{\max_{a \in A_q}{p_I(a|q_m)}}.
\end{align*}

The inference model may be any existing language model taken off-the-shelf (i.e. zero-shot) or finetuned on the task.
We do not do any further finetuning with knowledge prompting.
\section{Experimental Setup}
\label{sec:experiments}

Here, we describe the implementation details of our method and how they are adapted to each task. 

For knowledge generation, we use GPT-3 \cite{brown2020language} as the underlying language model, where our few-shot prompting method is most effective.
We generate $M=20$ knowledge statements for each question with nucleus sampling $p=0.5$ \cite{holtzman2019curious}, and discard repetitions and empty strings.
Generation is terminated when it exceeds 64 tokens or hits the $\texttt{\textbackslash n}$ token.\footnote{An exception is with the CSQA2 dataset, where for the best results we choose $M=5$ and allow for up to 128 tokens in each generation.}

For inference, we use off-the-shelf T5 \cite{raffel2019exploring} and GPT-3, as well as finetuned models that are state-of-the-art on each dataset, including UnifiedQA (UQA) \cite{khashabi-etal-2020-unifiedqa} and Unicorn \cite{lourie2021unicorn}.
See details in the task setup below.

\subsection{Datasets and Task Setup}

We evaluate our method on four commonsense reasoning datasets which cover a variety of challenges and problem formats.

\vspace{.1cm} \noindent \textbf{NumerSense} \cite{lin-etal-2020-birds} consists of numerical statements about common objects and concepts where for each sentence we need to recover a masked number word.
The choices are integers ranging from zero to ten, plus the word \textit{no}, so the task can be framed as a multiple-choice problem.
Since NumerSense is a diagnostic dataset, we only use zero-shot inference models, which is the current SOTA.
We follow \citet{stanford} who uses the state-of-the-art zero-shot T5 with text-infilling setup and select the choice with highest likelihood on its token(s).
We also implement zero-shot GPT-3 inference, where we plug in each choice to the question and compute the choice probability as the generative probability of the entire sentence, normalized over all the choices.

\vspace{.1cm} \noindent \textbf{CommonsenseQA (CSQA)} \cite{talmor-etal-2019-commonsenseqa} is a 5-way multiple-choice QA dataset about common world scenarios.
We do inference with the zero-shot and finetuned T5 models.
For zero-shot T5, we format the question as text-infilling, and predict the choice with highest sequence-to-sequence language modeling probability.
For finetuned T5 (including UnifiedQA which is SOTA), we use the same setup as \citet{khashabi-etal-2020-unifiedqa}.

\vspace{.1cm} \noindent \textbf{CommonsenseQA 2.0 (CSQA2)} \cite{talmor2021commonsenseqa} is a binary classification dataset where we need to judge whether commonsense statements are true or false.
We only do inference with the finetuned model, due to poor calibration of zero-shot models on this dataset.
We use finetuned Unicorn \cite{lourie2021unicorn}, which is the current SOTA, following the setup in \citet{talmor2021commonsenseqa}.

\vspace{.1cm} \noindent \textbf{QASC} \cite{khot2020qasc} is an 8-way multiple-choice QA dataset about grade school science.
This dataset also includes two pieces of background knowledge per question, whose composition fully answers the question.
We do inference with zero-shot T5 and finetuned T5 (including UnifiedQA which is SOTA), using the same setups as CSQA.

\subsection{Inference Model Setup}

Since all the inference models we use (T5, UnifiedQA, Unicorn) are generative language models, the support to a choice by the inference model is
\begin{align*}
& p_I(a | q) = \frac{\exp s_I(a | q)}{\sum_{a' \in A_q}{\exp s_I(a' | q)}}, \\
& \text{where } s_I(a | q) = \sum_{i=1}^{|a|}{\log{p(a_i | a_{<i}, q)}},
\end{align*}
and $a_i$ is the $i$-th token of choice $a$.

\subsection{Knowledge Generation Baselines}

We study the impact of our knowledge generation method (shorthanded as $K$) by comparing with the following baselines:

\vspace{-.2cm} \paragraph{No knowledge ($\varnothing$)} We refer to inference without any knowledge statements as the \textit{vanilla} baseline.

\vspace{-.2cm} \paragraph{Random sentences ($R$)} Sampling random sentences from the language model without conditioning on the question. We use the same implementation setup as our knowledge generation method (i.e. also using GPT-3, with the same hyperparameters).

\vspace{-.2cm} \paragraph{Context sentences ($C$)} Sampling sentences from the context of the question. This is implemented by sampling text continuations of the question from the language model. We use the same implementation setup as our knowledge generation method.

\vspace{-.2cm} \paragraph{Template-generated knowledge ($T$)} Self-talk \cite{shwartz-etal-2020-unsupervised} uses manually-designed templates to elicit knowledge statements from language models.
For fair comparison, we use GPT-3 as the knowledge generator in self-talk, and bound the number of generations to $M=20$ per question.
Templates and other hyperparameters are kept the same as their original paper.

\vspace{-.2cm} \paragraph{Retrieval-based knowledge ($\mathit{IR}$)} Instead of being generated, knowledge can be retrieved from appropriate sources. We consider the following retrieval-based methods. For NumerSense, knowledge is retrieved from sentences in Wikipedia and GenericsKB. For CSQA2, we use snippets returned by Google when querying the question. For QASC, we use the associated fact sentences that are used to create each question.

\vspace{-.2cm} \paragraph{Answers ($A$)} Instead of generating knowledge, GPT-3 can be prompted to generate direct answers to questions. In the prompts, we use the same input questions as those in knowledge generation, while replacing the knowledge statement with the ground truth answer. We consider two baselines: (1) Generate one answer per question and use this to measure the performance of the few-shot GPT-3 inference model; (2) Generate $M = 20$ answers per question, and use these answers to prompt the SOTA inference models.




\section{Experimental Results}

\begin{table*}[t]
\setlength{\tabcolsep}{3pt}
\centering
\resizebox{\textwidth}{!}{%
\begin{tabular}{c l uuu v w xx yy zz}
& & \multicolumn{3}{c}{$A$} & \multicolumn{1}{c}{$B_1$} & \multicolumn{1}{c}{$B_2$} & \multicolumn{2}{c}{$C$} & \multicolumn{2}{c}{$D_1$} & \multicolumn{2}{c}{$D_2$} \\
\toprule
& \textbf{Dataset} & \multicolumn{3}{c}{\textbf{NumerSense}} & \multicolumn{1}{c}{\textbf{CSQA}} & \multicolumn{1}{c}{\textbf{CSQA}} & \multicolumn{2}{c}{\textbf{CSQA2}} & \multicolumn{2}{c}{\textbf{QASC}} & \multicolumn{2}{c}{\textbf{QASC}} \\
& \textbf{Inference Model} & \multicolumn{3}{c}{T5-11b} & \multicolumn{1}{c}{T5-11b} & \multicolumn{1}{c}{UQA-11b-ft} & \multicolumn{2}{c}{Unicorn-ft} & \multicolumn{2}{c}{T5-11b} & \multicolumn{2}{c}{UQA-11b-ft} \\
& & dev & test$_{\text{core}}$ & test$_{\text{all}}$ & dev & dev & dev & test & dev & test & dev & test \\
\midrule
\multirow{6}{*}{\rotatebox[origin=c]{90}{\textbf{Knowledge Gen.}}} & ($\varnothing$) Vanilla baseline & 67.5 & 70.23 & 64.05 & 39.89 & 85.18 & 69.9 & 70.2$^{\dagger}$ & 48.16 & 44.89 & 81.75 & 76.74 \\
& ($R$) Random sentences & 68.5 & -- & -- & 21.79 & \textbf{85.42} & 70.37 & -- & 49.35 & -- & 82.18 & -- \\
& ($C$) Context sentences & \uline{70.5} & -- & -- & 42.51 & \uline{85.34} & 70.92 & -- & 55.83 & -- & 82.61 & -- \\
& ($T$) Template-based & -- & -- & -- & \uline{45.37} & -- & -- & -- & -- & -- & -- & -- \\
& ($\mathit{IR}$) Retrieval-based & -- & \uline{70.41} & \uline{65.10}$^{**}$ & -- & -- & \textbf{74.0} & \textbf{73.3}$^{\dagger\dagger}$ & \textbf{76.89} & -- & \textbf{90.06} & -- \\
& ($A$) Answers & 73.0 & -- & -- & 51.84 & 84.93 & 69.22 & -- & 52.48 & -- & 81.53 & -- \\
& ($K$) Ours & \textbf{78.0} & \textbf{79.24} & \textbf{72.47} & \textbf{47.26} & \uline{85.34} & \uline{72.37} & \uline{73.03} & \uline{58.32} & 55.00 & \uline{84.02} & 80.33 \\
\midrule
& prev. SOTA (no IR) & -- & 72.61 & 66.18$^*$ & -- & 79.1 (test)$^{\#}$ & 69.9 & 70.2$^{\dagger}$ & -- & -- & 81.75 & 76.74$^{\ddagger}$ \\
& Few-shot GPT-3 Infer. & 60.5 & -- & -- & -- & 71.58 & 53.80 & -- & -- & -- & 66.09 & -- \\
\bottomrule
\end{tabular}
}%
\caption{
    Experimental results of applying different knowledge generation methods on various tasks and inference models.
    T5-11b is the zero-shot inference model, whereas other inference models are finetuned based on T5-11b.
    We \textbf{bold} the best and \uline{underline} the second best numbers.
    Previous SOTA and retrieval-based methods are also based on the inference model in their corresponding column:
    * T5-11b 1.1 +digits (Submission by ISI Waltham);
    ** T5-11b + IR \cite{uscink};
    \# UQA-11b-ft \cite{khashabi-etal-2020-unifiedqa} (SOTA of single-model methods without referencing ConceptNet);
    $\dagger$ Unicorn-ft \cite{talmor2021commonsenseqa};
    $\dagger\dagger$ Unicorn-ft + Google snippets \cite{talmor2021commonsenseqa};
    $\ddagger$ UQA-11b-ft \cite{khashabi-etal-2020-unifiedqa}.
}
\label{tab:results}
\end{table*}

As we will show, our generated knowledge prompting method sets new state-of-the-art results on most datasets we evaluate on, and works well under both zero-shot and finetuned settings.
In particular, our knowledge generation outperforms naive baselines as well as template-based knowledge generation, and is on-par with retrieval-based systems.

\subsection{Overall Performance}

\autoref{tab:results} shows the results on zero-shot and finetuned models following our task setups.

\vspace{-.2cm} \paragraph{New state-of-the-art.}
We apply our method on top of the same inference model used in the previous state-of-the-art.
On NumerSense, we achieve a 6\% (66.18 $\rightarrow$ 72.47) improvement over the previous best method based on the zero-shot T5 model.
The previous state-of-the-art among non-retrieval methods on CSQA2 is based on the finetuned Unicorn model, upon which we improve by 2\% (70.2 $\rightarrow$ 73.03).
For QASC, the previous best is based on the finetuned UnifiedQA model, upon which we improve by 3\% (76.74 $\rightarrow$ 80.33).

\vspace{-.2cm} \paragraph{Zero-shot settings.} 
Columns $A$, $B_1$, and $D_1$ in \autoref{tab:results} show that our method substantially improves zero-shot inference models, by 7\% to 10\% across NumerSense (64.05 $\rightarrow$ 72.47), CSQA (39.89 $\rightarrow$ 47.26), and QASC (44.89 $\rightarrow$ 55.00).

\vspace{-.2cm} \paragraph{Finetuned settings.}
Columns $B_2$, $C$, and $D_2$ in \autoref{tab:results} indicate that our method consistently improves upon the vanilla baseline set by finetuned inference models (though by smaller margins than in the zero-shot settings).

\subsection{Knowledge Generation Methods}
\label{sec:quantitative}

\autoref{tab:results} reports the performance with different knowledge generation baselines.
Generally, random sentences barely help and even hurt the inference model, whereas context sentences of the question provide some gain.
In contrast, knowledge generated by our method consistently leads to substantial performance improvements, which implies that our knowledge is of high quality.

\vspace{-.2cm} \paragraph{Knowledge is an essential factor.}
The few-shot GPT-3 model is poorly calibrated to directly answer commonsense questions, underperforming our best models by 14\% to 20\% across all tasks.
Even when we use answers generated by few-shot GPT-3 to prompt the SOTA inference models, this still significantly falls behind our method on almost all the tasks and models we consider (with one exception -- CSQA with T5 inference).
Through the medium of \textit{knowledge}, our method can effectively leverage useful information possessed by GPT-3 to help improve even the SOTA models on various commonsense reasoning tasks.

\vspace{-.2cm} \paragraph{Our knowledge outperform template generated knowledge.}
We compare our knowledge generation method with the template-based \textit{self-talk} on the CSQA dev set.
(CSQA is the only task we experiment with that has self-talk templates available.)
Our method leads to a larger improvement over the T5-11b baseline than self-talk (by 1.89\%), showing that it is better at eliciting helpful knowledge from models. 

\vspace{-.2cm} \paragraph{Our knowledge is comparable with retrieval-based knowledge.}
On NumerSense, the retrieved knowledge only improves inference performance by 0.18\% on test-core and 1.02\% on test-all, while our method further outperforms it by 8.83\% and 7.37\%, respectively.
This shows that knowledge retrieved from a loosely-related knowledge base can be far less useful than our generated knowledge.
On CSQA2, although we are not able to beat the web-retrieved knowledge, our method still bridges the performance gap without referring to Google search.
For QASC, the ``retrieved'' knowledge is actually gold knowledge from a knowledge base that was used to construct the dataset. As a result, our generated knowledge falls significantly short of the retrieved knowledge.
In summary, our generated knowledge is roughly comparable with retrieved knowledge in terms of downstream performance, and is most valuable when there is no appropriate in-domain knowledge base to retrieve from.

\subsection{Analysis}
\label{sec:analysis}

\begin{figure}[t]
\centering
\includegraphics[width=\linewidth]{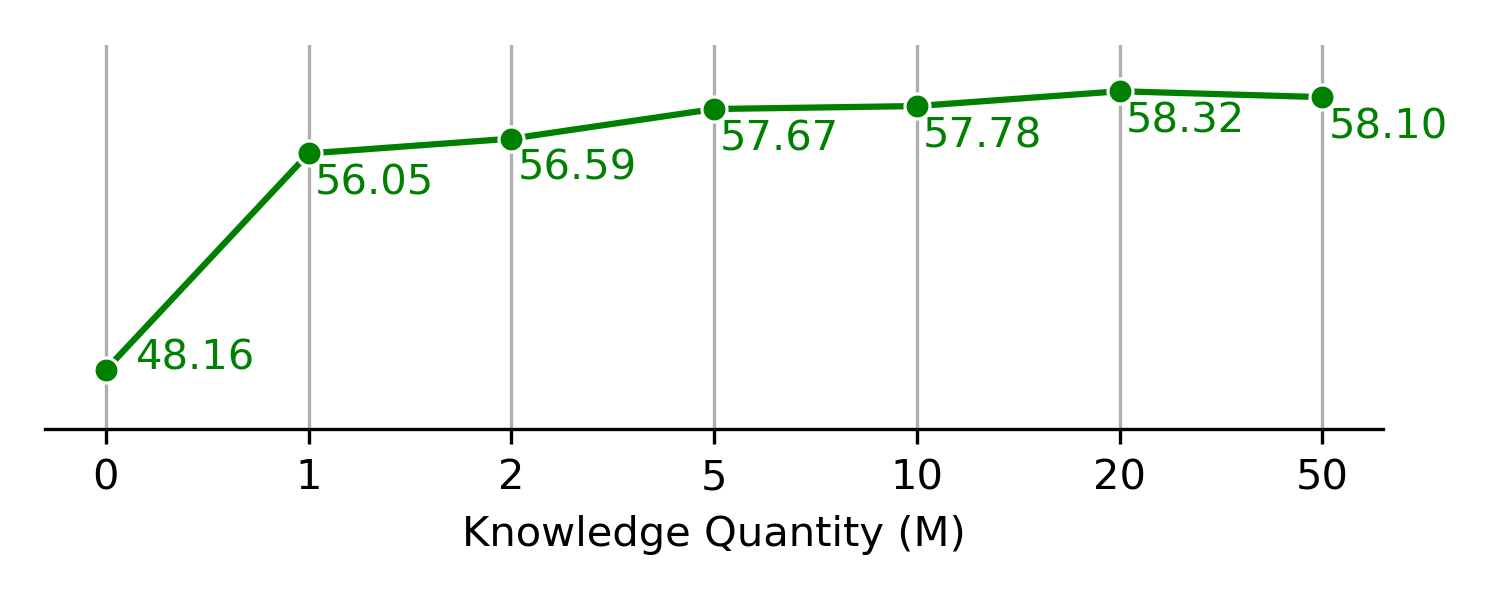}
\caption{
    Performance with different number of generated knowledge statements per question (QASC dev set, T5-11b inference model).
}
\label{fig:results_qasc_quantity}
\end{figure}

\paragraph{Better performance with more knowledge.}
We analyze the impact of the number of generated knowledge statements, $M$, and show the results in \autoref{fig:results_qasc_quantity}. 
Generally, the performance increases with the quantity of knowledge statements.
It saturates at $M=20$ and begins to decline when more knowledge statements are introduced, which may be because more noisy knowledge is generated. 

\begin{table}[t]
\footnotesize
\centering
\begin{tabular}{cc}
\toprule
\textbf{Integration method} & \textbf{QASC-dev} \\
\midrule
ours & \textbf{58.32} \\
Mixture-of-Experts & 56.26 \\
Product-of-Experts & 55.94 \\
\bottomrule
\end{tabular}
\caption{
    Performance with different knowledge integration methods
    (QASC dev set, T5-11b inference model).
}
\label{tab:results_qasc_ensembling}
\end{table}
\vspace{-.2cm} \paragraph{The knowledge integration method.}
In addition to the knowledge integration method described in \S\ref{sec:method_inference}, we experiment with two alternatives: Mixture-of-Experts (MoE) and Product-of-Experts (PoE) \cite{hinton2002training}.
These make the following modifications to \autoref{eqn:max_ensembling}, respectively:
\begin{align}
\text{MoE: } p_I(a|q,K_q) &\propto \sum_{0 \le m \le M}{p_I(a|q_m)}, \\
\text{PoE: } p_I(a|q,K_q) &\propto \prod_{0 \le m \le M}{p_I(a|q_m)}.
\end{align}
The results in \autoref{tab:results_qasc_ensembling} indicate that our knowledge integration method -- i.e. adaptively choosing the best knowledge to rely on -- is best among the three.

\begin{figure}[t]
\centering
\includegraphics[width=\linewidth]{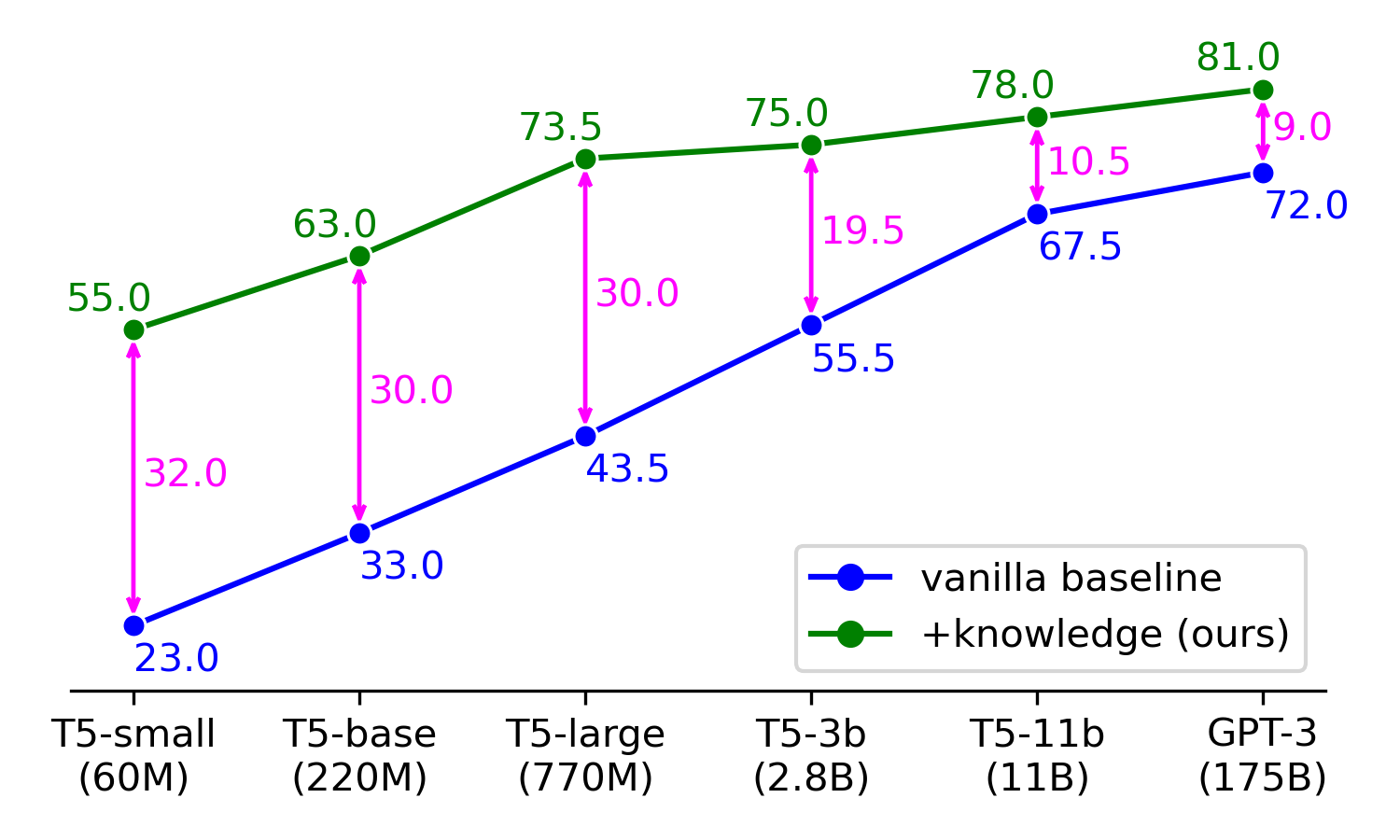}
\caption{
    Improvement on top of different sizes of inference model (Numersense dev set).
}
\label{fig:results_numersense_size}
\end{figure}

\begin{figure}[t]
\centering
\includegraphics[width=\linewidth]{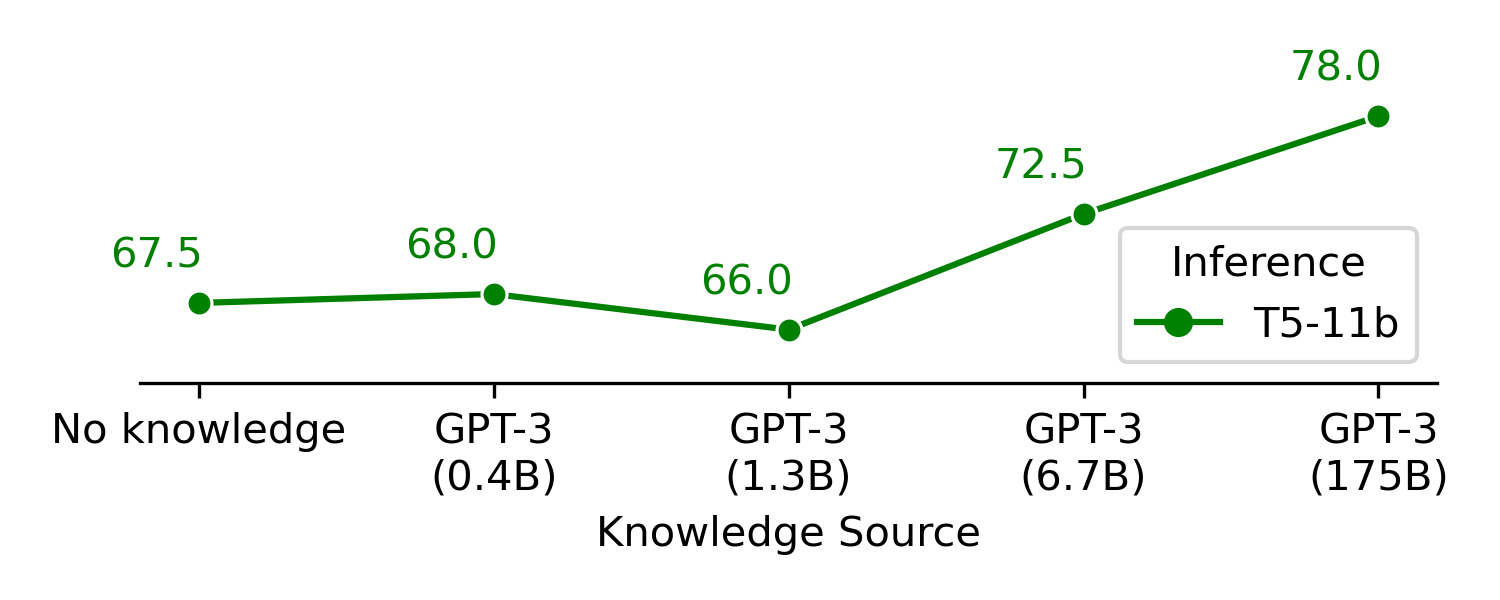}
\caption{
    Improvement by different sizes of knowledge generation model (Numersense dev set, T5-11b inference model).
}
\label{fig:results_numersense_knowledge}
\end{figure}
\vspace{-.2cm} \paragraph{Lightweight inference models and amplification.}
We found that the size of inference model affects the magnitude of improvement.
\autoref{fig:results_numersense_size} shows the NumerSense performance gain on top of different sizes of inference model.
As we use smaller inference models, the performance gain increases drastically.
In particular, with our method the smallest T5 model is as powerful as the T5-3b baseline, and T5-large outperforms the GPT-3 baseline.
This indicates that model-generated knowledge can enable high performing, yet lightweight, inference models.
Furthermore, the improvement does not diminish as the inference model becomes as big as the knowledge generation model, as the inference by GPT-3 can benefit by 9.0\% from the knowledge elicited from itself.
This indicates that our method can somewhat \textit{amplify} the useful knowledge already possessed by the model, leading to better predictions.

\vspace{-.2cm} \paragraph{The size of knowledge generation model.}
\autoref{fig:results_numersense_knowledge} shows the NumerSense performance gain when using different sizes of GPT-3 as the knowledge generation model.
On top of the T5-11b inference model, The 6.7B knowledge model gives a 5.0\% improvement, narrower than the 10.5\% improvement given by the 175B knowledge model.
The 1.3B and 0.4B knowledge models do not give a significant improvement.
Therefore, we do not necessarily need the largest version of GPT-3 as the knowledge source, though we do need the model to be relatively large in order to generate useful and reliable knowledge.

\subsection{Human Evaluation}
\label{sec:human}

\begin{figure*}[t]
\begin{minipage}{.4\linewidth}
\centering
\includegraphics[width=\linewidth]{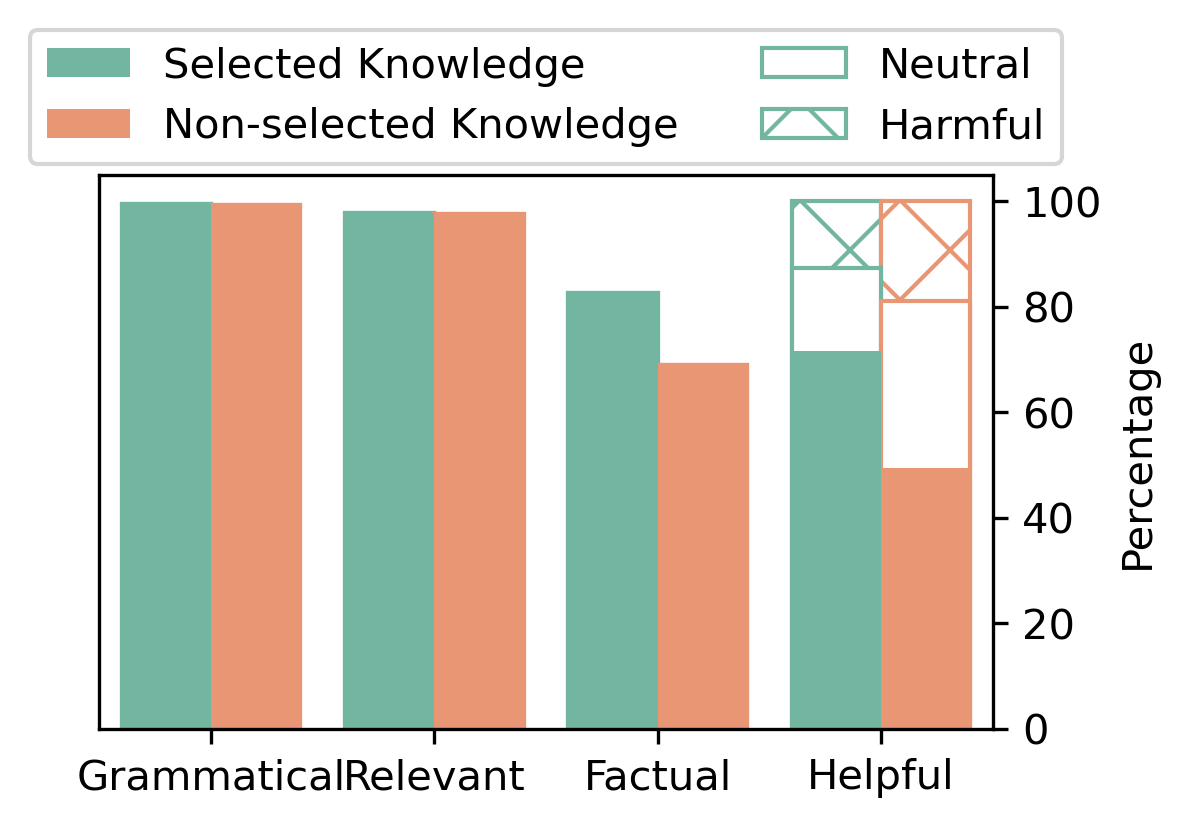}
\end{minipage}
\hfill
\begin{minipage}{.58\linewidth}
\centering
\includegraphics[width=\linewidth]{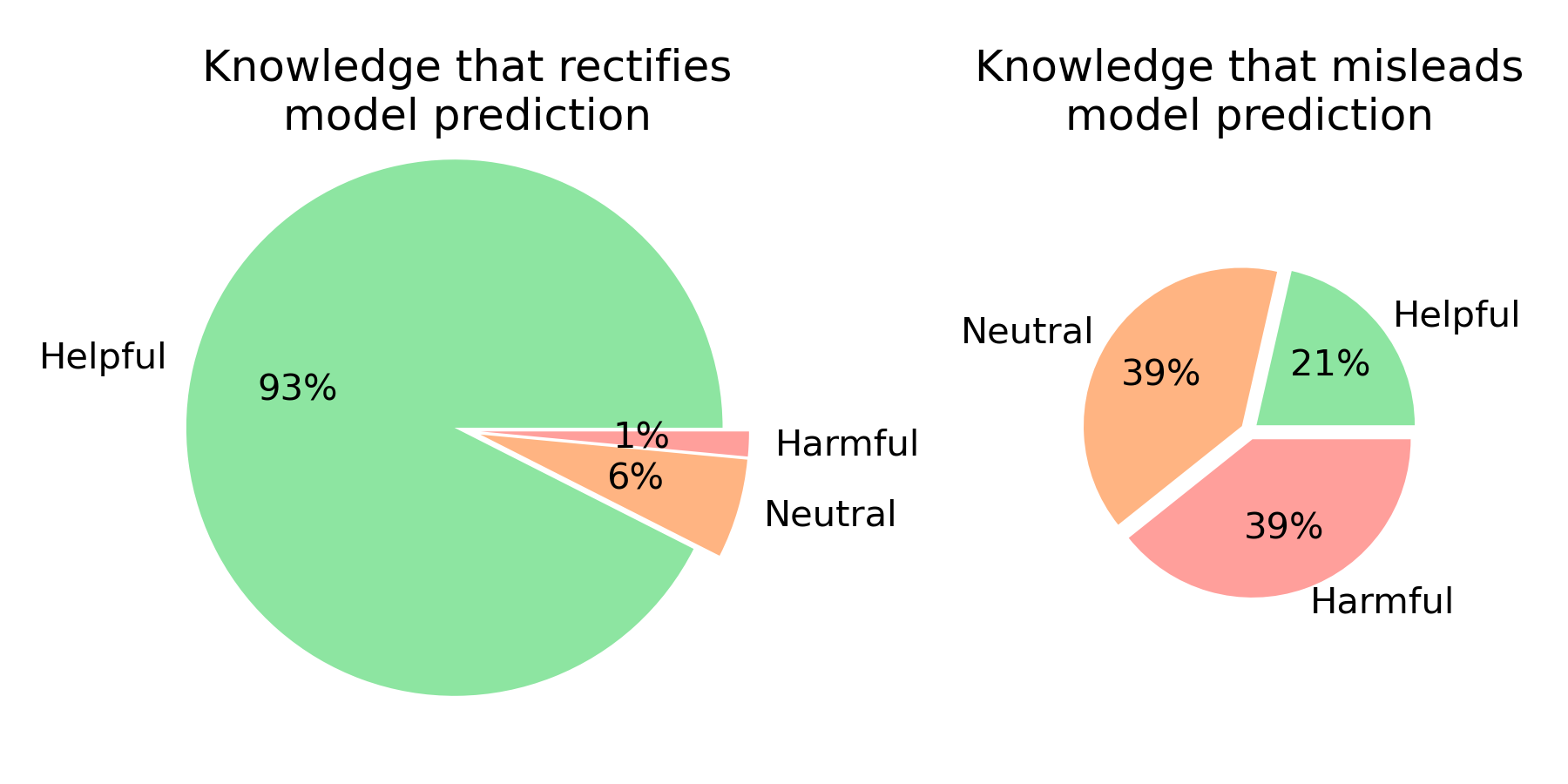}
\end{minipage}
\caption{
    Human evaluation of generated knowledge.
    \textbf{Left:} Percentage of good knowledge statements along each axis.
    \textbf{Right:} Agreement between human and machine on helpfulness of selected knowledge.
}
\label{fig:human}
\end{figure*}

We conduct a human evaluation on NumerSense and QASC to study the quality of generated knowledge and the interpretability of its impact on task performance. 

\vspace{-.2cm} \paragraph{Evaluation.} We report the quality of knowledge statements along four axes: (1) \textit{Grammaticality}: whether it is grammatical; (2) \textit{Relevance}: whether it is relevant to the topic or concepts mentioned on the question; (3) \textit{Factuality}: whether it is (mostly) factually correct; and (4) \textit{Helpfulness}: whether it helps answering the question in an either direct or indirect way, and may fall into one of the three categories: helpful (i.e. supports the correct answer), harmful (i.e. negates the correct answer or supports an incorrect answer), or neutral (neither helpful nor harmful).
These metrics are adapted from \citet{shwartz-etal-2020-unsupervised} and are defined in Appendix \ref{sec:guidelines}.

From each dataset, we sample up to 50 \textit{selected knowledge} (\S\ref{sec:method_inference}) that change the correctness of T5-11b's prediction (i.e. rectifies model prediction from wrong to right, or misleads model prediction from right to wrong).
The knowledge are labeled by two NLP experts and a moderate level of agreement was reached (Fleiss Kappa $\kappa = 0.57$ \cite{landis1977measurement}).
To ensure objectivity, it is not revealed to the annotators whether the knowledge rectifies or misleads the model prediction.

\vspace{-.2cm}  \paragraph{Results.} \autoref{fig:human} summarizes the results.
The vast majority of selected knowledge are grammatical and relevant to the question, and 83\% of them are factually correct.
72\% are seen as being helpful for answering the question according the human evaluators, whereas 13\% are harmful.
Out of the knowledge statements that rectify the model predictions, 93\% are labeled as helpful by the human evaluators; in contrast, when the knowledge statement misleads the model, only 21\% are labeled as helpful, and 39\% harmful.
Of the knowledge deemed helpful by human \textit{and} rectifies model prediction, 95\% are factual, while of those deemed harmful by human \textit{and} misleads model prediction, 86\% are non-factual, suggesting that improving knowledge factuality is a promising path towards more helpful knowledge.
We also analyzed the non-selected knowledge and found that these statements have slightly lower factuality and helpfulness than the selected knowledge.

\subsection{Qualitative Examples}
\label{sec:qualitative}

\begin{table*}[t]
\footnotesize
\centering
\begin{tabular}{cp{7cm}ccc}
\toprule
\textbf{Dataset} & \textbf{Question} / \textsl{Knowledge} & \textbf{Prediction} & \textbf{Score} & \textbf{Reasoning} \\
\midrule
\multirow{2}{*}{NumerSense} & \textbf{clams have evolved to have [M] shells.} & \textcolor{wrong}{no} & \textcolor{wrong}{0.37} | \textcolor{right}{0.18} & \textcolor{gray}{Commonsense} \\
& \qquad \textsl{Clams have a bivalve shell.} & \textcolor{right}{two} & \textcolor{right}{0.89} & Paraphrasing \\
\midrule
\multirow{2}{*}{NumerSense} & \textbf{an easel can have [M] or four legs.} & \textcolor{wrong}{two} & \textcolor{wrong}{0.45} | \textcolor{right}{0.45} & \textcolor{gray}{Commonsense} \\
& \qquad \textsl{A tripod is a kind of easel.} & \textcolor{right}{three} & \textcolor{right}{0.46} & Induction \\
\midrule
\multirow{2}{*}{CSQA} & \textbf{Where does a heifer's master live?} & \textcolor{wrong}{slaughter house} & \textcolor{wrong}{0.89} | \textcolor{right}{0.01} & \textcolor{gray}{Commonsense} \\
& \qquad \textsl{The master of a heifer is a farmer.} & \textcolor{right}{farm house} & \textcolor{right}{0.92} & Deduction \\
\midrule
\multirow{2}{*}{CSQA} & \textbf{Aside from water and nourishment what does your dog need?} & \textcolor{wrong}{walked} & \textcolor{wrong}{0.55} | \textcolor{right}{0.04} & \textcolor{gray}{Commonsense} \\
& \qquad \textsl{Dogs need attention and affection.} & \textcolor{right}{lots of attention} & \textcolor{right}{0.91} & Elimination \\
\midrule
\multirow{2}{*}{CSQA} & \textbf{I did not need a servant. I was not a what?} & \textcolor{wrong}{in charge} & \textcolor{wrong}{0.47} | \textcolor{right}{0.32} & \textcolor{gray}{Commonsense} \\
& \qquad \textit{People who have servants are rich.} & \textcolor{right}{rich person} & \textcolor{right}{0.99} & Abduction \\
\midrule
\multirow{2}{*}{CSQA2} & \textbf{Part of golf is trying to get a higher point total than others.} & \textcolor{wrong}{yes} & \textcolor{wrong}{1.00} | \textcolor{right}{0.00} & \textcolor{gray}{Commonsense} \\
& \qquad \textit{The player with the lowest score wins.} & \textcolor{right}{no} & \textcolor{right}{1.00} & Negation \\
\midrule
\multirow{2}{*}{CSQA2} & \textbf{Eighth plus eight is smaller than fifteen.} & \textcolor{wrong}{yes} & \textcolor{wrong}{0.97} | \textcolor{right}{0.03} & \textcolor{gray}{Commonsense} \\
& \qquad \textit{Eighth plus eight is sixteen, which is larger than fifteen.} & \textcolor{right}{no} & \textcolor{right}{1.00} & Numerical \\
\midrule
\multirow{2}{*}{QASC} & \textbf{[M] is used for transportation.} & \textcolor{wrong}{plastic} & \textcolor{wrong}{0.41} | \textcolor{right}{0.12} & \textcolor{gray}{Commonsense} \\
& \qquad \textit{Bicycles are used for transportation.} & \textcolor{right}{boats} & \textcolor{right}{0.74} & Analogy \\
\bottomrule
\end{tabular}
\caption{
    More examples where prompting with generated knowledge reduces the reasoning type and rectifies the prediction.
    The first row of each section is the original question and the inference results associated with it; the second row is a model-generated knowledge statement that prompts the inference model.
    We show \textcolor{right}{correct answers in green}, \textcolor{wrong}{incorrect answers in red}, and their corresponding scores assigned by the inference model.
}
\label{tab:qualitative}
\end{table*}

\autoref{tab:qualitative} shows a few examples where the generated knowledge rectifies model prediction.
Due to space constraints we only show the \textit{selected knowledge} (\S\ref{sec:method_inference}) for each question.
In all examples, the model without prompted knowledge assigns a higher score to an incorrect answer than the correct answer, while with knowledge prompting, the correct answer is assigned a much higher score. 
Prompting with generated knowledge can transform commonsense reasoning into explicit reasoning procedures such as paraphrasing, induction, deduction, analogy, abductive reasoning, logical elimination, negation, and numerical reasoning.
\section{Related Work}

\paragraph{Knowledge can be elicited from pretrained language models.}
Numerous works have shown that pretrained language models implicitly contain a large amount of knowledge that can be queried via conditional generation \cite{davison-etal-2019-commonsense, petroni-etal-2019-language, jiang-etal-2020-know}.
Consequently, these models can directly perform inference on tasks like commonsense reasoning \cite{trinh2018simple, yang-etal-2020-designing}, text classification \cite{shin-etal-2020-autoprompt, puri2019zero}, and natural language inference \cite{shin-etal-2020-autoprompt, schick-schutze-2021-exploiting}.
Inspired by these observations, we elicit question-related knowledge in an explicit form from language models and use them to guide the inference.

\vspace{-.2cm} \paragraph{Leveraging external knowledge for commonsense reasoning.}
Some work uses external commonsense knowledge bases to make improvements on various NLP tasks, including commonsense reasoning.
One approach is to inject commonsense knowledge into language models, either by pretraining on knowledge bases \cite{ma2021knowledge, chang-etal-2020-incorporating, mitra2019additional, zhong2019improving} or finetuning the model so that it can reason with additional retrieved knowledge \cite{chang-etal-2020-incorporating, mitra2019additional, bian2021benchmarking}.
Another direction is to ground the question into a knowledge graph and do inference with graph-based reasoning \cite{lin-etal-2019-kagnet, lv2020graph, yasunaga-etal-2021-qa}.

A common prerequisite of these methods is a high-quality, high-coverage, in-domain commonsense knowledge base \cite{ma-etal-2019-towards}.
Some commonsense reasoning datasets are derived from existing knowledge bases; for example, CommonsenseQA \cite{talmor-etal-2019-commonsenseqa} is derived from ConceptNet \cite{speer2017conceptnet}, and Social IQA \cite{sap-etal-2019-social} is derived from ATOMIC \cite{sap2019atomic}. 
For such datasets, it is natural to elicit related knowledge from the underlying knowledge base that derived them, and typically this would demonstrate considerable gains \cite{mitra2019additional, chang-etal-2020-incorporating}.
However, if there is a domain mismatch between the dataset and the knowledge base, such gains tend to diminish \cite{mitra2019additional, ma-etal-2019-towards}.
This becomes a bottleneck when encountering datasets that have no suitable knowledge base (e.g. NumerSense \cite{lin-etal-2020-birds} and CommonsenseQA 2.0 \cite{talmor2021commonsenseqa}), or when the system needs to handle commonsense queries that do not fit in any of the commonsense domains represented by an existing knowledge base.
Our work overcomes this difficulty by leveraging  pretrained language models as the source of commonsense knowledge.

\vspace{-.2cm} \paragraph{Adding generated text during inference.}
Recently, several works show that model performance on commonsense reasoning can be boosted by augmenting the question with model-generated text, such as clarifications, explanations, and implications.
Self-talk \cite{shwartz-etal-2020-unsupervised} elicits clarifications to concepts in the question and appends them to the inference model input.
Contrastive explanations \cite{paranjape-etal-2021-prompting} prompts inference models with generated explanations that contrast between two answer choices.
The aforementioned methods depend on task-specific templates to inquire the generator, which means they are only capable of eliciting a limited variety of knowledge and require careful hand-crafting to transfer to new tasks.
Other explanation-based methods \cite{latcinnik2020explaining, rajani-etal-2019-explain} finetune the generator model so that it produces explanations that are used for question augmentation.
DynaGen \cite{bosselut2021dynamic} uses pretrained commonsense models to generate implications of a question and builds a dynamic graph of natural language statements on which reasoning is conducted.
However, its usage of COMeT \cite{bosselut-etal-2019-comet} as the generator confines its applicability to the social commonsense domain.
Our work contributes to this general line of research, yet different from these previous methods that elicit knowledge with task-specific templates or from finetuned knowledge generators, 
our method requires only a few human-written demonstrations in the style of the task, making it much more flexible, easy-to-transfer, and engineering-efficient.
\section{Conclusion}


We introduce generated knowledge prompting, a simple method to elicit and integrate knowledge from language models so as to improve performance on commonsense reasoning tasks.
In particular, we generate knowledge statements by prompting a language model with task-specific, human-written, few-shot demonstrations of question-knowledge pairs.
We show that knowledge can be integrated by simply plugging it in at inference time, with no need to finetune the model for knowledge integration. 
Our method shows effectiveness across multiple datasets, sets the new state-of-the-art on three commonsense reasoning tasks, and works under a variety of settings.
The method's success highlights language models as sources of flexible, high-quality knowledge for commonsense reasoning.

\section*{Acknowledgements}

This work was funded in part by the Natural Sciences and Engineering Research Council of Canada (NSERC) (funding reference number 401233309), DARPA MCS program through NIWC Pacific (N66001-19-2-4031), and the Allen Institute for AI. We also thank Google Cloud Compute, as well as OpenAI.

We thank Daniel Khashabi, Vered Shwartz, Bhargavi Paranjape, Bill Yuchen Lin, Jonathan Herzig for their help with the experiments and evaluation.

\bibliography{anthology,custom}
\bibliographystyle{acl_natbib}

\clearpage

\appendix
\section{Appendix}
\label{sec:appendix}

\subsection{Comparison with Prior Methods}
\begin{table*}[b]
\footnotesize
\centering
\begin{tabular}{ccc}
\toprule
\textbf{Method} & \textbf{Knowledge Generator} & \textbf{Inference Model} \\
\midrule
\textbf{CAGE} \cite{rajani-etal-2019-explain} & task-finetuned & joint-finetuned \\
\citet{latcinnik2020explaining} & task-finetuned & joint-finetuned \\
\textbf{DynaGen} \cite{bosselut2021dynamic} & task-finetuned & joint-finetuned \\
\textbf{Self-talk} \cite{shwartz-etal-2020-unsupervised} & template-prompted & 0-shot \\
\textbf{Contrastive expl.} \cite{paranjape-etal-2021-prompting} & template-prompted & 0-shot \& joint-finetuned \\
\midrule
\textbf{Generated knowledge prompting} (ours) & demonstrations-prompted & 0-shot \& task-finetuned \\
\bottomrule
\end{tabular}
\caption{
    Comparison of methods that add generated text to an inference model.
    \textbf{Knowledge Generator}: \textit{task-finetuned} -- a model finetuned to generate task-specific knowledge; \textit{template-prompted} -- an off-the-shelf LM from which knowledge statements are elicited via templates; \textit{demonstration-prompted} -- an off-the-shelf LM from which knowledge statements are elicited via few-shot demonstrations (\S\ref{sec:method_generation}).
    \textbf{Inference Model}: \textit{0-shot} -- an off-the-shelf LM that is set up to make predictions; \textit{task-finetuned} -- a model finetuned with task training data (and without seeing extra knowledge); \textit{joint-finetuned} -- a model finetuned with task training data \textit{and} generated knowledge.
}
\label{tab:methods}
\end{table*}

\autoref{tab:methods} summarizes the comparison between our generated knowledge prompting method and prior methods that add generated text to an inference model for commonsense reasoning tasks.
Our method is unique because it uses few-shot demonstrations to prompt for knowledge generation, and can apply to finetuned inference models without joint finetuning with knowledge.

\subsection{Prompts for Knowledge Generation}
\label{sec:prompts}

\begin{table*}[h!]
\footnotesize
\centering
\begin{tabular}{r|p{10cm}}
\textbf{Task} & \textbf{Prompt} \\ 
NumerSense & Generate some numerical facts about objects. Examples:\\
\\
& Input: \textbf{penguins have <mask> wings.} \\ 
& Knowledge: \textit{Birds have two wings. Penguin is a kind of bird.} \\ 
\\
& Input: \textbf{a parallelogram has <mask> sides.} \\ 
& Knowledge: \textit{A rectangular is a parallelogram. A square is a parallelogram.} \\ 
\\
& Input: \textbf{there are <mask> feet in a yard.} \\ 
& Knowledge: \textit{A yard is three feet.} \\ 
\\
& Input: \textbf{water can exist in <mask> states.} \\ 
& Knowledge: \textit{There states for matter are solid, liquid, and gas.} \\ 
\\
& Input: \textbf{a typical human being has <mask> limbs.} \\ 
& Knowledge: \textit{Human has two arms and two legs.} \\ 
\\
& Input: \textbf{\{question\}}\\
& Knowledge:
\end{tabular}
\caption{
    Prompt for knowledge generation on NumerSense.
    Demonstration examples are manually written and the knowledge enables explicit reasoning procedures to answer the input question. 
}
\label{tab:prompt_numersense}
\end{table*}
\begin{table*}[h!]
\footnotesize
\centering
\begin{tabular}{r|p{12cm}}
\textbf{Task} & \textbf{Prompt} \\ 
CSQA & Generate some knowledge about the concepts in the input. Examples:\\
\\
& Input: \textbf{Google Maps and other highway and street GPS services have replaced what?} \\ 
& Knowledge: \textit{Electronic maps are the modern version of paper atlas.} \\ 
\\
& Input: \textbf{The fox walked from the city into the forest, what was it looking for?} \\ 
& Knowledge: \textit{Natural habitats are usually away from cities.} \\ 
\\
& Input: \textbf{You can share files with someone if you have a connection to a what?} \\ 
& Knowledge: \textit{Files can be shared over the Internet.} \\ 
\\
& Input: \textbf{Too many people want exotic snakes.  The demand is driving what to carry them?} \\ 
& Knowledge: \textit{Some people raise snakes as pets.} \\ 
\\
& Input: \textbf{The body guard was good at his duties, he made the person who hired him what?} \\ 
& Knowledge: \textit{The job of body guards is to ensure the safety and security of the employer.} \\ 
\\
& Input: \textbf{\{question\}}\\
& Knowledge:
\end{tabular}
\caption{
    Prompt for knowledge generation on CSQA.
    Demonstration examples are selected from the CSQA training set; we manually write relevant knowledge for each input question.
}
\label{tab:prompt_csqa}
\end{table*}
\begin{table*}[h]
\footnotesize
\centering
\begin{tabular}{r|p{12cm}}
\textbf{Task} & \textbf{Prompt} \\
CSQA2 & Generate some knowledge about the input. Examples:\\
\\
& Input: \textbf{Greece is larger than mexico.}\\
& Knowledge: \textit{Greece is approximately 131,957 sq km, while Mexico is approximately 1,964,375 sq km, making Mexico 1,389\% larger than Greece.}\\
\\
& Input: \textbf{Glasses always fog up.}\\
& Knowledge: \textit{Condensation occurs on eyeglass lenses when water vapor from your sweat, breath, and ambient humidity lands on a cold surface, cools, and then changes into tiny drops of liquid, forming a film that you see as fog. Your lenses will be relatively cool compared to your breath, especially when the outside air is cold.}\\
\\
& Input: \textbf{A fish is capable of thinking.}\\
& Knowledge: \textit{Fish are more intelligent than they appear. In many areas, such as memory, their cognitive powers match or exceed those of 'higher' vertebrates including non-human primates. Fish's long-term memories help them keep track of complex social relationships.}\\
\\
& Input: \textbf{A common effect of smoking lots of cigarettes in one's lifetime is a higher than normal chance of getting lung cancer.}\\
& Knowledge: \textit{Those who consistently averaged less than one cigarette per day over their lifetime had nine times the risk of dying from lung cancer than never smokers. Among people who smoked between one and 10 cigarettes per day, the risk of dying from lung cancer was nearly 12 times higher than that of never smokers.}\\
\\
& Input: \textbf{A rock is the same size as a pebble.}\\
& Knowledge: \textit{A pebble is a clast of rock with a particle size of 4 to 64 millimetres based on the Udden-Wentworth scale of sedimentology. Pebbles are generally considered larger than granules (2 to 4 millimetres diameter) and smaller than cobbles (64 to 256 millimetres diameter).}\\
\\
& Input: \textbf{\{question\}}\\
& Knowledge:
\end{tabular}
\caption{
    Prompt for knowledge generation on CSQA2.
    Demonstration examples are selected from the CSQA2 training set; we use the annotated \textit{Google featured snippet} as the knowledge.
}
\label{tab:prompt_csqa2}
\end{table*}
\begin{table*}[h]
\footnotesize
\centering
\begin{tabular}{r|p{12cm}}
\textbf{Task} & \textbf{Prompt} \\ 
QASC & Generate some knowledge about the input. Examples:\\
\\
& Input: \textbf{What type of water formation is formed by clouds?} \\ 
& Knowledge: \textit{Clouds are made of water vapor.} \\ 
\\
& Input: \textbf{What can prevent food spoilage?} \\ 
& Knowledge: \textit{Dehydrating food is used for preserving food.} \\ 
\\
& Input: \textbf{The process by which genes are passed is} \\ 
& Knowledge: \textit{Genes are passed from parent to offspring.} \\ 
\\
& Input: \textbf{The stomach does what in the body?} \\ 
& Knowledge: \textit{The stomach is part of the digestive system.} \\ 
\\
& Input: \textbf{What can cause rocks to break down?} \\ 
& Knowledge: \textit{Mechanical weathering is when rocks are broken down by mechanical means.} \\ 
\\
& Input: \textbf{\{question\}}\\
& Knowledge:
\end{tabular}
\caption{
    Prompt for knowledge generation on QASC.
    Demonstration examples are selected from the QASC training set; we use one of the gold separate facts as the knowledge.
}
\label{tab:prompt_qasc}
\end{table*}

\autoref{tab:prompt_numersense} through \ref{tab:prompt_qasc} shows the full prompts for knowledge generation that we use for each evaluated task: NumerSense, CSQA, CSQA2, and QASC.

\subsection{Human Evaluation Guidelines}
\label{sec:guidelines}

\begin{table*}
\footnotesize
\centering
\begin{tabular}{l p{2.5cm} p{10cm}}
\toprule
\textbf{Attribute} & \textbf{Options} & \textbf{Description and Examples} \\
\midrule
\textbf{Grammaticality} & grammarical; ungrammatical but understandable; completely gibberish & Whether the knowledge statement is grammatical. We are aware that some of the statements are not fully grammatical. If you can still understand what the statement says, even if it's an incomplete sentence or slightly ungrammatical, please select the "ungrammatical but understandable" option. \\
\midrule
\textbf{Relevance} & relevant; not relevant & Whether a knowledge statement is relevant to the given question. A statement is relevant if it covers one the same topics as the question, or contains a salient concept that is same or similar to one in the question. Examples: \\
\\
& & \textcolor{red}{[Question]} you may take the subway back and forth to work <mask> days a week. \\
& & \textcolor{orange}{[Knowledge]} You take the subway back and forth to work five days a week. \\
& & \textcolor{blue}{[Judgment]} Relevant, because the question and knowledge are both about the topic of business days. \\
\\
& & \textcolor{red}{[Question]} a bradypus torquatus is native to brazil and has <mask> toes on each limb. \\
& & \textcolor{orange}{[Knowledge]} A bradypus torquatus is a kind of mammal. A mammal has four limbs. \\
& & \textcolor{blue}{[Judgment]} Relevant, because the question and knowledge share a common salient concept "bradypus torquatus". \\
\midrule
\textbf{Factuality} & factual; not factual & Whether a knowledge statement is (mostly) factually correct or not. If there are exceptions or corner cases, it can still be considered factual if they are rare or unlikely. Examples: \\
\\
& & \textcolor{orange}{[Knowledge]} A limousine has four doors. \\
& & \textcolor{blue}{[Judgment]} Factual. \\
\\
& & \textcolor{orange}{[Knowledge]} A human hand has four fingers and a thumb. \\
& & \textcolor{blue}{[Judgment]} Factual, despite that there are exceptions -- people with disabilities may have less or more fingers. \\
\\
& & \textcolor{orange}{[Knowledge]} A rectangle is a shape with two equal sides. \\
& & \textcolor{blue}{[Judgment]} Not factual, because a rectangle has four sides. \\
\midrule
\end{tabular}
\caption{Human evaluation guidelines. Continued in \autoref{tab:guidelines_2}.}
\label{tab:guidelines}
\end{table*}

\begin{table*}
\footnotesize
\centering
\begin{tabular}{l p{2.5cm} p{10cm}}
\toprule
\textbf{Attribute} & \textbf{Options} & \textbf{Description and Examples} \\
\midrule
\textbf{Helpfulness} & helpful; neutral; harmful & Whether a knowledge statement provides useful information in support OR contradiction of the answer. A statement is helpful if it supports the answer either directly or indirectly. More on indirect support -- The statement might not directly answer the question directly, yet it may support an indirect reasoning path that reaches the answer. A statement is harmful if it negates the answer or supports an alternative potential answer either directly or indirectly. A statement is neutral if it is neither helpful nor harmful. Examples: \\
\\
& & \textcolor{red}{[Question]} you may take the subway back and forth to work <mask> days a week. \\
& & \textcolor{green}{[Answer]} five \\
& & \textcolor{orange}{[Knowledge]} You take the subway back and forth to work five days a week. \\
& & \textcolor{blue}{[Judgment]} Helpful. Because the statement directly supports the answer. \\
\\
& & \textcolor{red}{[Question]} spiders have <mask> legs. \\
& & \textcolor{green}{[Answer]} eight \\
& & \textcolor{orange}{[Knowledge]} Arachnids have eight legs. \\
& & \textcolor{blue}{[Judgment]} Helpful. Although the statement does not directly refer to spiders, together with the fact that "spiders are a kind of arachnids" it completes a reasoning chain in deriving the answer. \\
\\
& & \textcolor{red}{[Question]} a game of chess may have <mask> outcomes. \\
& & \textcolor{green}{[Answer]} three \\
& & \textcolor{orange}{[Knowledge]} A game of chess has two outcomes. \\
& & \textcolor{blue}{[Judgment]} Harmful. Since the statement supports answering "two" instead of "three". \\
\\
& & \textcolor{red}{[Question]} a bradypus torquatus is native to brazil and has <mask> toes on each limb. \\
& & \textcolor{green}{[Answer]} three \\
& & \textcolor{orange}{[Knowledge]} A bradypus torquatus is a kind of mammal. A mammal has four limbs. \\
& & \textcolor{blue}{[Judgment]} Neutral. The statement does not provide information in favor or contrast of the answer. \\
\bottomrule
\end{tabular}
\caption{(continued) Human evaluation guidelines.}
\label{tab:guidelines_2}
\end{table*}

\autoref{tab:guidelines} and \ref{tab:guidelines_2} shows the detailed guidelines we use for human evaluation of generated knowledge.

\section{Checklist}
\label{sec:checklist}

\subsection{Limitations and Risks}

\paragraph{Limitations.}
Our method is tested on a representative selection of commonsense reasoning tasks and datasets.
Applying this method to other tasks may require people with moderate expertise to craft a task-specific prompt to feed into the method.

\paragraph{Risks.}
It is possible that our proposed method may lower the performance of commonsense reasoning systems, if not implemented properly or using badly-designed prompts.
Such risk can be mitigated by following the prompt design guidelines in this paper (\S\ref{sec:method_generation}).

\subsection{Computation}

We do not train any new model in this paper.
Inference is conducted on Quadro RTX 8000 GPUs and costs about 200 GPU hours in total.
Knowledge generation is done with the OpenAI GPT-3 API, with an approximate cost of \$500.

Our method is implemented with PyTorch and the Huggingface Transformers library.

\end{document}